\pdfoutput=1

\documentclass[11pt]{article}

\usepackage[final]{acl2024}

\usepackage{times}
\usepackage{latexsym}
\usepackage{pifont}
\newcommand{\cmark}{\text{\ding{51}}}
\newcommand{\xmark}{\text{\ding{55}}}

\usepackage[T1]{fontenc}

\usepackage[utf8]{inputenc}

\usepackage{microtype}

\usepackage{inconsolata}

\usepackage{amsfonts}
\usepackage{amsmath}
\usepackage{booktabs}
\usepackage{multirow}
\usepackage{subcaption}

\usepackage{graphicx}
\usepackage{algorithm}
\usepackage{algpseudocode}
\algrenewcommand\algorithmicrequire{\textbf{Input:}}
\algrenewcommand\algorithmicensure{\textbf{Output:}}

\usepackage{caption}
\usepackage{bbold}
\usepackage{cleveref}
\crefname{section}{§}{§§}

\usepackage{tikz}
\newcommand\mybox[2][]{\tikz[overlay]\node[fill=blue!20,inner sep=2pt, anchor=text, rectangle, rounded corners=1mm,#1] {#2};\phantom{#2}}

\usepackage{color,soul}
\usepackage{paralist}
\usepackage[shortlabels]{enumitem}

\definecolor{fam}{rgb}{1.0, 0.0, 0.00}
\definecolor{orig}{rgb}{0.1, 1.0, 0.1}

\title{On the Influence of Gender and Race in \\Romantic Relationship Prediction from Large Language Models}

\author{Abhilasha Sancheti$^*$~~~~~~~Haozhe An$^*$~~~~~~~Rachel Rudinger \\
    University of Maryland, College Park \\
    \texttt{\{sancheti, haozhe, rudinger\}@umd.edu} \\
}

\begin{document}
\maketitle
\begin{abstract}
We study the presence of heteronormative biases and prejudice against interracial romantic relationships in large language models by performing controlled name-replacement experiments for the task of relationship prediction. 
We show that models are less likely to predict romantic relationships for \begin{inparaenum}[(a)]
\item{same-gender character pairs than different-gender pairs; and}
\item{intra/inter-racial character pairs involving Asian names as compared to Black, Hispanic, or White names.
We examine the contextualized embeddings of first names and find that gender for Asian names is less discernible than non-Asian names.
We discuss the social implications of our findings, underlining the need to prioritize the development of inclusive and equitable technology.
}
\end{inparaenum}
\end{abstract}


\def\thefootnote{*}\footnotetext{These authors contributed equally to this work.}
\def\thefootnote{\arabic{footnote}}

\section{Introduction}

\label{sec:intro}

Identifying romantic relationships from a given dialogue presents a challenging task in natural language understanding~\cite{jia2021ddrel, tigunova-etal-2021-pride}. 
The perceived gender, race, or ethnicity of the speakers, often inferred from their names, may inadvertently lead a model to predict a relationship type that conforms to conventional societal views.
We hypothesize that, when predicting romantic relationships, models may mirror \textit{heteronormative biases}~\cite{pollitt2021heteronormativity,vasquez-etal-2022-heterocorpus} and prejudice against interracial romantic relationships~\cite{lewandowski2001, miller2004} present in humans and society. Heteronormative biases assume and favor traditional gender roles, heterosexual relationships, and nuclear families, often marginalizing other gender expressions, sexuality, and family dynamics. In the US, legal protections for interracial and gay marriages were not achieved nationwide until 1967 and 2015, respectively. These relationships continue to face prejudice and discrimination in the present days~\cite{buist2019lgbtq, knauer2020lgbtq, zambelli2023, pittman2024, shaji2024}.
 
\begin{figure}[t]
		\centering
            \includegraphics[width=0.98\linewidth]{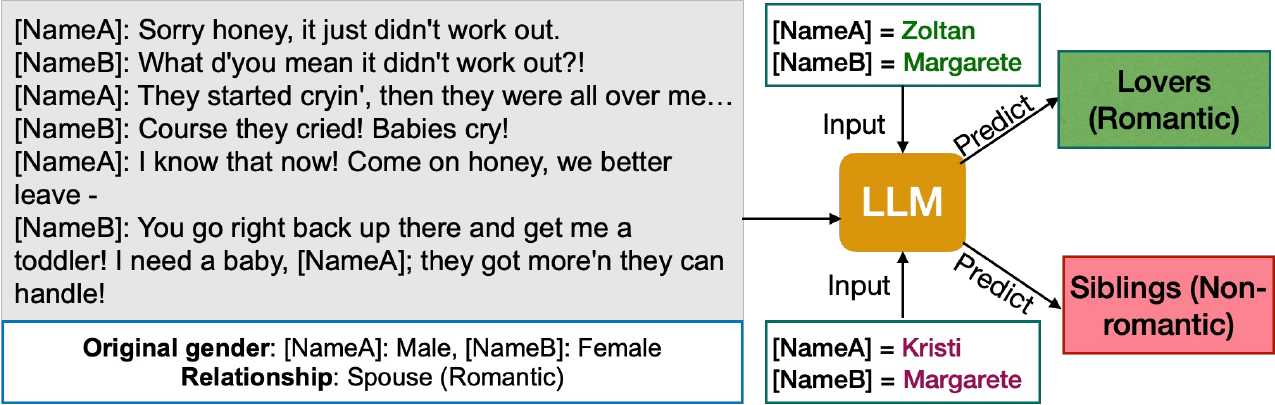}
	\caption{Sample conversation from DDRel~\citep{jia2021ddrel} dataset and relationships predicted by Llama2-7B when characters are replaced by names with \mybox[fill=orig!50]{different-gender} and \mybox[fill=fam!30]{same-gender}. LLM tends to predict differently despite the same conversation.\protect
 }
	\label{fig:task}
\end{figure}

In this paper, we consider the task of predicting romantic relationships from dialogues in movie scripts to study
\textit{whether LLMs make such predictions based on the demographic attributes associated with a pair of character names, 
in ways that 
reflect heteronormative biases and prejudice against interracial romantic relationships}.
For instance, Figure~\ref{fig:task} shows a conversation between a female and a male spouse pair, for which
Llama2-7B predicts a romantic relationship when the names in the conversation are replaced with a pair of different-gender names, but predicts a non-romantic relationship when replaced by same-gender names.

Ideally, name-replacement should not significantly alter the predictions of a fair and robust model, as the utterance content plays a more substantial role in language understanding, despite the potential interdependence between utterances and original names.
Different predictions suggest that a model may be prone to overlooking romantic relationships that diverge from societal norms, 
thus raising ethical concerns.
Such behavior would indicate that language models inadequately represent certain societal groups~\cite{blodgett-etal-2020-language}, potentially exacerbating stigma surrounding relationships~\cite{rosenthal2015relationship, reczek2020sexual} and sidelining underrepresented groups~\cite{nozza-etal-2022-measuring, felkner-etal-2023-winoqueer}.


Through controlled character name-replacement experiments, 
we find that relationships between
\begin{inparaenum}[(a)]
\item{same-gender character pairs; and}
\item{intra/inter-racial character pairs involving Asian names} are less likely to be predicted as romantic.
\end{inparaenum}
These findings reveal how some LLMs may stereotypically interpret interactions between people, potentially reducing the recognition of non-mainstream relationship types. While prior work studies gender and racial biases by identifying stereotypical attributes of \textit{individuals}~\cite{cao-etal-2022-theory, cheng-etal-2023-marked, an-etal-2023-sodapop}, this paper investigates the role of gender and race in LLMs' inferences about \textit{relationships} between \textit{two} individuals using a relationship prediction dataset~\cite{jia2021ddrel}.

\section{Experimental Setup}
\label{sec:setup}
We define the following task.
Given a conversation $C$ which consists of a sequence of turns $\left( (S_1, u_1), (S_2, u_2), \dots, (S_n, u_n)\right)$ between characters $A$ and $B$, where $S_i \in \{S_A, S_B\}$ indicates that the speaker of an utterance ($u_i$, $i \in \{1:n\}$) is either $A$ or $B$, the task 
is to identify the relationship represented as a categorical label from a pre-defined set.
We carry out controlled name-replacement experiments by prompting LLMs (zero-shot) to predict the relationship type between $A$ and $B$ given $C$.
\paragraph{Models} We study Llama2 (\{7B, 13B\}-chat)~\citep{touvron2023llama} with its official implementation,\footnote{\url{https://github.com/facebookresearch/llama}} and Mistral-7B-Instruct~\citep{jiang2023mistral} using its huggingface implementation. Hyperparameters are specified in~\textsection\ref{sec:models}.
%

\paragraph{Dataset} We use the test set of DDRel~\cite{jia2021ddrel} which consists of movie scripts from IMSDb, with annotations for relationship labels between the characters according to $13$ pre-defined types (Table~\ref{tab:relationship-dist} in appendix). 
We consider Lovers, Spouse, or Courtship predictions as romantic and the rest as non-romantic. 
For our experiments, we use $327$ instances of the test set in which characters originally have different genders (manually annotated) because the test set has no dialogues between same-gender characters with the romantic label. We discuss the limitations of this study due to data source representation issues at the end of this paper.

\paragraph{Prompt Selection} As LLMs are sensitive to prompts~\citep{min-etal-2022-rethinking}, we experimented with several prompt formulations on the original data (test set) for accuracy, and selected the prompt (see Figure~\ref{fig:prompts} in appendix) resulting in the highest accuracy which was closest to scores reported by others~\cite{jia2021ddrel, ou2024dialogbench}. We note that our prompt selection is done prior to running the name-replacement experiments.

\paragraph{Evaluation} We compare the average recall of predicting romantic relationships across different gender assignments and races/ethnicities. We study recall as we hypothesize heteronormative and interracial relationship biases would manifest as low (romantic) recall for same-gender and interracial groups. For completeness, we also report the mean precision, F1, and accuracy scores in~\textsection\ref{sec:additional}.

\subsection{Studying the Influence of Gender Pairings} 

\label{sec:gender-experiments}
\label{sec:same-gender} 

\begin{figure*}[t]
	\centering
		\centering
            \includegraphics[width=0.98\linewidth]{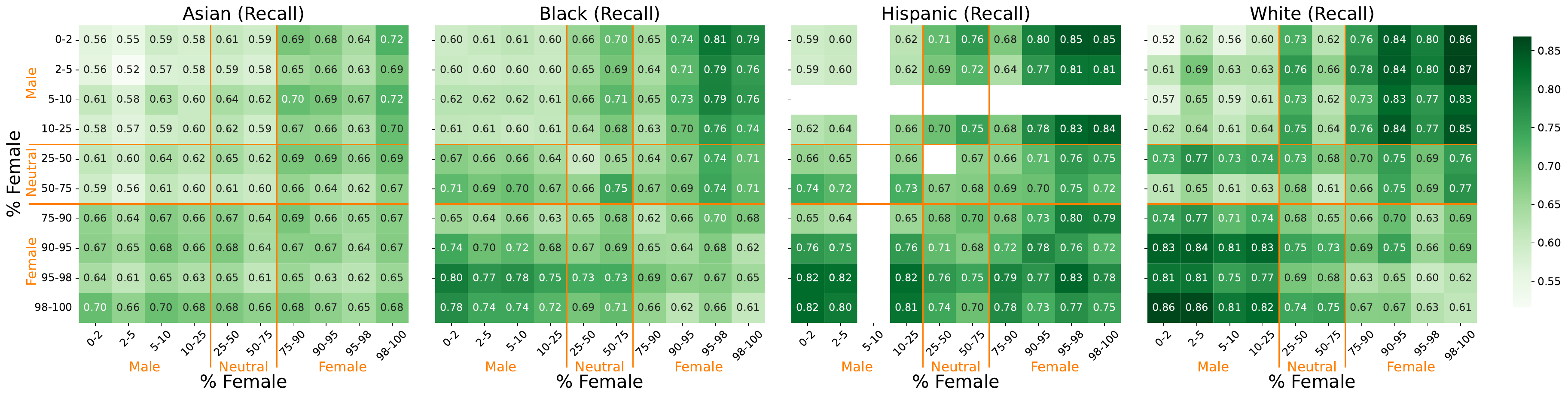}
	\caption{Recall of predicting romantic relationships from Llama2-7B for subset of the dataset where characters originally have different genders. Horizontal and vertical axes denote \% female of the name replacing an originally female and male character name from the dialogue.
 The upper-triangle (lower-triangle) shows the scores when names are replaced preserving (swapping) the genders of characters' names as-is in the original conversation. 
  We consider the names with lesser \% female as male names for determining gender preservation for name-replacement. 
 }
	\label{fig:gender-spectrum}
\end{figure*}

We ask whether the models are equally likely to recognize romantic relationships for character pairs of varying gender assignments and if this behavior is the same across different races. We hypothesize that models are prone to heteronormative bias and are more likely to predict romantic relationships for contrastive gender assignments.
To test this, we collect $30$ names per race,\footnote{Except for Hispanic wherein we did not get any names in $5-10\%$ bin and only 1 name in $25-50\%$ bin.} dividing them into $10$ non-linearly segmented bins that cover gender-neutral names (shown in Figure~\ref{fig:gender-spectrum}) based on the percentage of population that has been assigned as female at birth. Detailed name inclusion criteria and data sources are elaborated in~\textsection\ref{sec:gender_spectrum_names}. We replace the original name-pair in each conversation with all pairs of distinct names per race. 

As dialogues may reveal gender identities (\textit{e.g.}, ``sir'', ``ma'am'', ``father'', etc.), we manually identify a
subset ($271$ instances) where such explicit cues are absent (to the best of our judgement) to minimize gender information leakage and avoid explicit gender inconsistency between the dialogue and the gender associated with the replaced name. 
In these dialogues, gendered pronouns typically refer to a third person who is not part of the conversation. As a result, they do not reveal the speakers' gender identity. However, pronouns can indicate the sexual orientation of a speaker (\textit{e.g.}, ``Betty: \textit{You do love him, don't you?}''). Such cues, along with other implicit cues about gender identity that are harder to detect, may confound our analysis. However, our findings as discussed in \textsection\ref{sec:findings} reveal that implicit cues are not a major confounding factor.
We discuss this aspect further in the Limitations section.

\subsection{Studying Intra/Inter-Racial Pairings}

\label{sec:race}
We examine whether the models exhibit prejudice against interracial romantic relationships when making predictions. We collect another set of $80$ first names that are both strongly race- and gender-indicative, evenly distributed among four races/ethnicities and two genders (details described in~\textsection\ref{sec:race_spectrum_names}). 
We perform pairwise name-replacements using these $80$ names for the $327$ test samples to analyze the relationship predictions among different intra/inter-racial name pairs.

We defer details related to full prompt used and model output parsing to~\textsection\ref{sec:exp-details}.



\section{Findings} \label{sec:findings}
\paragraph{Same-gender relationships are less likely to be predicted as romantic than different-gender ones.} 
We observe a significant variation in recall of romantic relationship predictions from Llama2-7B (see Figure~\ref{fig:gender-spectrum}) for name-replacements involving different (top-right, and bottom-left)- versus same-gender pairs. This reveals that the model conservatively predicts romantic relationships when both the characters have names associated with the same gender (top-left -- both male; bottom-right -- both female). However, the precision across all races ranges between $0.78-0.84$ (see Figure~\ref{fig:gender-spectrum-llama} in appendix). Such (relatively) low difference indicates that, while the model makes precise romantic predictions across all gender assignments and races, romantic predictions are more likely for contrastive gender assignments.
Higher recall (Figure~\ref{fig:gender-spectrum}) for both female (bottom-right) replacements than both male (top-left) across all races indicates a potential \textbf{stronger heteronormative bias against both male than both female pairs}. This could potentially be an effect of associating female names with romantic relationships as indicated by higher recall for female-neutral than male-neutral pairs. To test this hypothesis, we substitute one speaker's name with a male, female or neutral name while keeping the other anonymized (substituting with ``X''). We find that name pairs containing one female name tend to have higher recall than those containing one male name (Table~\ref{tab:anonymous-one-gender-results} in appendix). This could either be due to a stronger association of female names with romantic relationships in general, or stronger heteronormative bias against male-male romantic relationships if models are (effectively) marginalizing probabilities over the anonymous character. A possible explanation for the former is that women tend to be portrayed only as objects of romance in fictional works, \textit{e.g.}, as popularly evidenced by the failure of many movies to pass the Bechdel test~\citep{agarwal-etal-2015-key}.

The smaller gap in the recall between both female (bottom-right) name-replacements and different-gender (top-right and bottom-left) ones for Asian and Hispanic as compared to White and Black may result from model's inability to discern gender from Asian and Hispanic names as accurately as for White and Black names. Figures~\ref{fig:gender-spectrum-Llama2-13b} and~\ref{fig:gender-spectrum-mistral} (appendix) show similar trends for Llama2-13B and Mistral-7B, respectively.
\paragraph{The unnaturalness of movie scripts with name and gender substitutions could, in theory, provide an alternative explanation for the observed biases, but the evidence shows this is not the cause.}
As female characters may speak differently from male characters,
our name-replacements can introduce statistical inconsistency between the gender associated with a character name and the style or content of the lines they speak, potentially confounding our observations. However, comparable recall between
name-replacements that preserve the gender (upper-triangle; specifically top-right) associated with the original speakers and the swapped variants (lower-triangle; specifically bottom-left) in Figure~\ref{fig:gender-spectrum}, indicates that swapping both characters' genders has minimal impact on model's performance in the conversations we used. Hence, we conclude the potential inconsistency between gender and linguistic content is not a major confounding factor.

\begin{figure}[t]
	\centering
		\centering
    \includegraphics[width=0.75\linewidth]{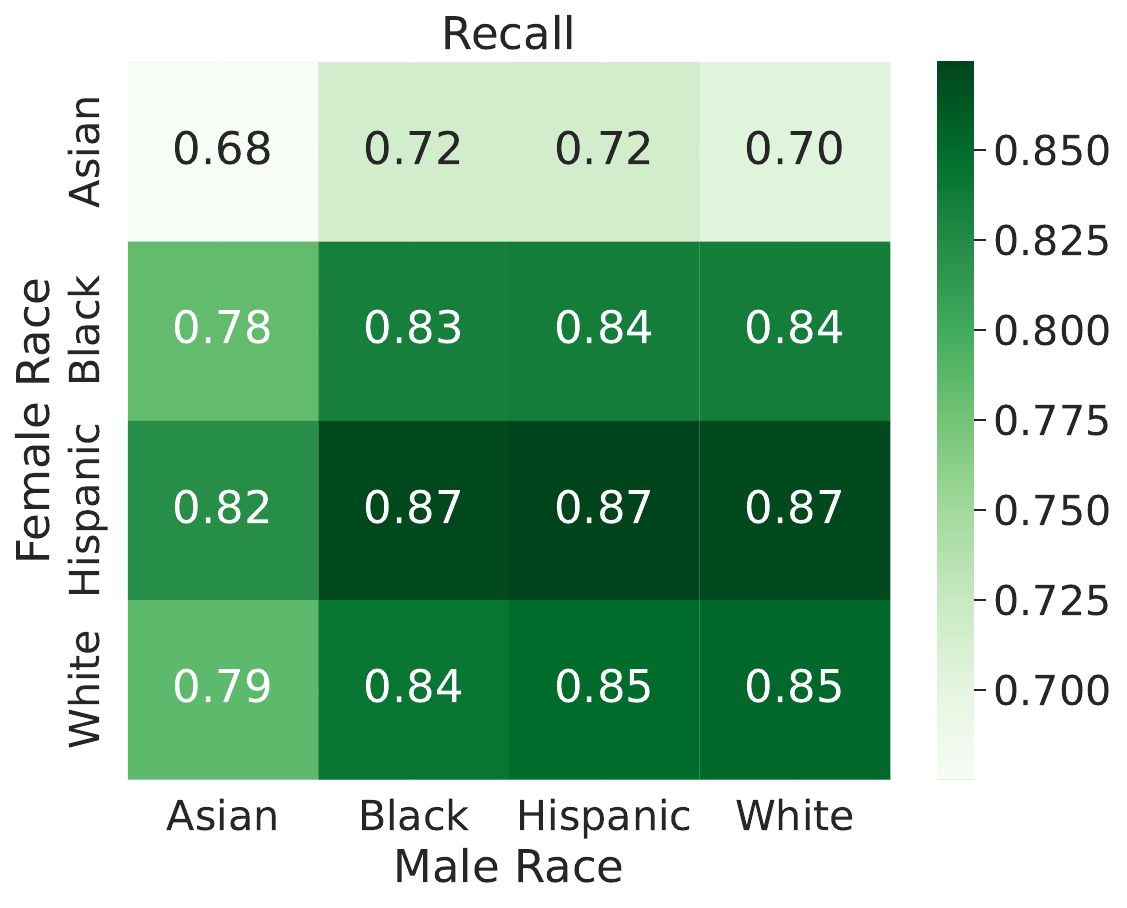}
	\caption{Recall of predicting romantic relationships from Llama2-7B for subset of the dataset where characters have different genders and are replaced with names associated with different races/ethnicities.}
	\label{fig:race-spectrum-llama7b}
\end{figure}

\paragraph{Character pairs involving Asian names have lower romantic recall; however, we do not find strong evidence against interracial pairings.}
While Llama2-7B has similar precision of predicting a romantic relationship across all racial pairs ($0.80$ -- $0.82$, shown in Figure~\ref{fig:race-spectrum-llama7b_appendix} in appendix), Figure~\ref{fig:race-spectrum-llama7b} shows name pairs involving at least one Asian name have significantly lower recall. Noticeably, the recall is the lowest ($0.68$) when both character names are associated with Asian.
Although there are variations in recall values among different racial setups, we do not observe disparate differences between interracial and intraracial name pairs for non-Asian names.
Results for Llama2-13B and Mistral-7B, shown respectively in Figure~\ref{fig:race-spectrum-llama13b} and~\ref{fig:race-spectrum-mistral7b} in the appendix, demonstrate a similar trend that Asian names lead to substantially lower recall values.
Such systematically worse performance on Asian names potentially perpetuates known algorithmic biases~\cite{chander2016racist, Akter2021, Papakyriakopoulos2023}.


\section{Analysis and Discussion} 
\label{sec:analysis}
We perform additional experiments to understand the observed model behavior.

\paragraph{Why does a model tend to predict fewer romantic relationships for racial pairings that involve Asian names?}
Although we select names for each race that have strong real-world statistical associations with one gender, we hypothesize that low recall on pairs with one or more Asian names may be due to model's inability to discern gender from Asian names. To test this hypothesis, we retrieve the contextualized embeddings from Llama2-7B for each first name (collected in \textsection\ref{sec:race}) occurrence in $15$ romantic and $15$ non-romantic random dialogues. We obtain $209,800$ embeddings, which are used to train logistic regression models that classify the gender or race associated with a name (details in~\textsection\ref{sec:exp-details}).
As we compare the average classification accuracy (across 5 different train-test splits) against a majority baseline, we observe, in Table~\ref{tab:embs_logistic_regression}, that gender could be effectively predicted for non-Asian name embeddings, and the embeddings are distinguishable by race for all races/ethnicities in a One-vs-All setting.
However, Asian name embeddings encode minimal gender information, decreasing the likelihood of a model leveraging the inferred gender identity when making relationship predictions that reflect heteronormative biases.

\begin{table}[]
    \centering
    \resizebox{\linewidth}{!}{ 
    \begin{tabular}{@{}ll|llll@{}}
        \toprule
        \multicolumn{2}{l|}{Race/Ethnicity}                               & \multicolumn{1}{c}{Asian} & \multicolumn{1}{c}{Black} & \multicolumn{1}{c}{Hispanic} & \multicolumn{1}{c}{White} \\ \midrule
        \multicolumn{1}{l|}{\multirow{2}{*}{Gender}} & Logistic regression & 53.3$\pm$12.7           & \textbf{96.4$\pm$2.9}            & \textbf{80.5$\pm$13.0}              & \textbf{99.9$\pm$0.2}            \\
        \multicolumn{1}{l|}{}                             & Majority baseline     & 54.2$\pm$0.0            & 54.2$\pm$0.0            & 54.2$\pm$0.0               & 53.9$\pm$0.3            \\ \midrule
        \multicolumn{1}{l|}{\multirow{2}{*}{Race}}   & Logistic regression & \textbf{97.6$\pm$1.9}            & \textbf{70.5$\pm$6.3}            & \textbf{89.5$\pm$4.1}               & \textbf{94.2$\pm$3.8}            \\
        \multicolumn{1}{l|}{}                             & Majority baseline    & 50.6$\pm$0.2            & 50.6$\pm$0.4            & 50.9$\pm$0.4               & 50.9$\pm$0.3            \\ \bottomrule
    \end{tabular}
    }
    \caption{Logistic regression classification accuracy (\%) of predicting the demographic attributes associated with a name from Llama2-7B contextualized embeddings.}
    \label{tab:embs_logistic_regression}
\end{table}

%

\begin{table}[]
    \centering
    \resizebox{\linewidth}{!}{ 
    \begin{tabular}{@{}l|cccc@{}}
        \toprule
        \textbf{Model} & \multicolumn{1}{c}{\textbf{Precision}} & \multicolumn{1}{c}{\textbf{Recall}} & \multicolumn{1}{c}{\textbf{F1}} & \multicolumn{1}{c}{\textbf{Accuracy}} \\ 
        \midrule
        \multicolumn{5}{c}{\multirow{1}{*}{\textbf{Gender Pairings}}} \\
        \midrule
        Llama2-7B & $0.7978$ & $0.6887$  & $0.7392$    & $0.6125$\\
        Llama2-13B & $0.8649$ & $0.3019$  & $0.4476$    & $0.4170$\\
        Mistral-7B & $0.8269$ & $0.2028$  & $0.3258$    & $0.3432$\\
        \midrule
         \multicolumn{5}{c}{\multirow{1}{*}{\textbf{Racial Pairings}}} \\
        \midrule
        Llama2-7B & $0.8063$ & $0.7131$  & $0.7569$    & $0.6422$\\
        Llama2-13B & $0.8696$ & $0.3287$  & $0.4665$    & $0.4404$\\
        Mistral-7B & $0.8406$ & $0.2311$  & $0.3625$    & $0.3761$\\
        \bottomrule
    \end{tabular}
    }
    \caption{Evaluation scores for anonymous name-replacements (character replaced with ``X'' or ``Y'') for different models under study. These results depict the model's performance solely based on the context.} \label{tab:anonymous-gender-results}
\end{table}
\paragraph{Does gender association have a stronger influence on model's prediction than race/ethnicity?}

We hypothesize that models' tendency to associate gender with names influences their relationship predictions. To test this, we substitute names with generic placeholders (``X'' and ``Y'') to get a baseline where a model has no access to character names (more details in~\textsection\ref{sec:anonymous-exp-details}). 
After name-replacements, any deviation from these results (Table~\ref{tab:anonymous-gender-results}) would indicate that a model exploits the implicit information from first names.
In Figure~\ref{fig:gender-spectrum}, multiple settings have recall values that significantly differ from those in the anonymized setting ($0.6887$). 
This disparity suggests name-replacements introduce gender information that significantly influences the model behavior.
Such trends are less prominent for Asian names due to the model's apparent inability to distinguish gender information in Asian names (Table~\ref{tab:embs_logistic_regression}).
By contrast, racial information encoded in first names exerts a lesser impact.
Non-Asian heterosexual intra/inter-racial pairs give rise to similar recall in Figure~\ref{fig:race-spectrum-llama7b}.
We thus do not observe strong prejudice against interracial romantic relationships here. 
\section{Social Implications}


It has been a prolonged and arduous struggle to recognize and accept gay marriages in the US~\cite{andersen2016transformative, duberman2019stonewall}.
Legal recognition of these relationships remains a challenge in many other countries~\cite{lee2017measuring, chia2019lgbtq, ramdas2021negotiating}.
Even within the US, LGBTQIA+ 
people still encounter discrimination~\cite{buist2019lgbtq, knauer2020lgbtq, naylor2020social}.

We believe heteronormative biases we have observed could impact various downstream LLM use cases, potentially causing both representational and allocational harms~\cite{blodgett-etal-2020-language}. For example, when LLMs are used for story generation based on social media posts as the premise~\cite{te-etal-2018-using, li2024pretrained}, the life events of members of the LGBTQIA+ community may be overlooked or misrepresented. If LLMs struggle to recognize same-gender romantic relationships, they may further marginalize the LGBTQIA+ community by diminishing their social visibility and representation.
In addition, such model behavior may result in uneven allocation of resources or opportunities.
Consider an online advertising system that promotes low-interest home loans for married couples based on social media interactions. A model unable to identify same-gender marriages would exclude these couples from the promotion. Therefore, building inclusive technology that respects minority rights is essential.

\section{Related Work}
Prior works~\cite{wang-etal-2022-measuring, jeoung-etal-2023-examining, sandoval-etal-2023-rose,wan-etal-2023-kelly, an-etal-2023-sodapop, an-etal-2024-large, nghiem2024you} show that language models often treat first names differently, even with controlled input contexts, due to factors like frequency and demographic attributes associated with names~\cite{hall-maudslay-etal-2019-name, shwartz-etal-2020-grounded, wolfe-caliskan-2021-low, czarnowska-etal-2021-quantifying, an-rudinger-2023-nichelle}.
Our work uses models' interpretations of gender associated with first names to reveal heteronormative biases in some LLMs.

Further, NLP systems often fail in interpreting various social factors (\textit{e.g.}, social norms, cultures, and relations) of language~\citep{hovy-yang-2021-importance}.  One such factor of interest is the representation of social relationships in these systems, including power dynamics~\citep{prabhakaran-etal-2012-predicting}, friendship~\citep{krishnan-eisenstein-2015-youre}, and romantic relationships~\citep{seraj2021language}.
Recently,~\citet{stewart-mihalcea-2024-whose} show failure of popular machine translation systems in translating sentences concerning relationships between nouns of same-gender.
Leveraging the task of relationship prediction and using an existing dataset~\cite{jia2021ddrel}, our work contributes to the assessment of social relationship-related biases in LLMs arising from gender and race associations with first names.

\section{Conclusion}
Through controlled name-replacement experiments, we find that LLMs predict romantic relationships between characters based on the demographic identities associated with their first names.
Specifically, relationship predictions between same-gender and intra/inter-racial character pairs involving Asian names are less likely to be romantic.
Our analysis of contextualized name embeddings sheds light on the cause of our findings.
We also highlight the social implications of this potentially harmful model behavior for the LGBTQIA+ community. We urge advocates to build technology that respects the rights of marginalized social groups.

\section*{Limitations} \label{sec:limitations}
\paragraph{Prompt sensitivity and in-context learning.} LLMs are sensitive to prompt formats~\citep{min-etal-2022-rethinking, li2024pedantspreciseevaluationsdiverse} therefore the accuracy of predictions may vary within or across models. While we had experimented with several prompts before converging to the one we use (gave the best prediction accuracy on the original dataset as well as close to that reported in \citet{jia2021ddrel}), future work may investigate the impact of different prompt formulations and if in-context learning can help in reducing the influence of biases on the downstream tasks.

\paragraph{Inadequate coverage of names associated with different identities.} 
We recognize that our paper has limitations regarding the number of races and genders studied. This is due to the unavailability of data sources to compile a sufficiently large number of names strongly associated with a wide range of underrepresented races and gender identities. 

\paragraph{Linguistic usage might be significantly different in same-gender romantic relationships.}
The test set we have utilized~\cite{jia2021ddrel} does not contain dialogues between same-gender character pairs in romantic relationships. As a consequence, we lack conversations that effectively depict interactions between same-gender partners. We acknowledge this limitation in our data source. 
However, in cases where same-gender partners exhibit behavior similar to different-gender couples, our results indicate that LLMs tend to demonstrate heteronormative biases in the intersection of these interaction styles.

\paragraph{Conversations might contain implicit gender-revealing cues.}
While we ensure consistency between gender associated with an utterance (based on how a male speaks vs a female) and the gender associated with a name by only considering the conversations that do not have explicit gender-revealing cues as described in \textsection{\ref{sec:gender-experiments}}, we acknowledge the possibility of the presence of implicit gender-revealing cues which is harder to detect. However, we believe that our findings stand valid even if the implicit cues are present as demonstrated by comparable recall between name-replacements that preserve the gender (upper-triangle; specifically top-right) associated with the original speaker and the swapped variants (lower-triangle; specifically bottom-left) in Figure~\ref{fig:gender-spectrum}. We leave further analysis of the nuances with implicit cues to future work.

\section*{Ethical Considerations}
\paragraph{Inconsistency between self-identification and demographic attributes associated with a name.}
Our categorization of names into subgroups of race/ethnicity and gender is based on real-world data as we observe a strong statistical association between names and demographic attributes (race/ethnicity and gender). 
However, it is crucial to realize that a person with a particular name may identify themselves differently from the majority, 
and we should respect their individual preferences and embrace the differences. 
We have attempted to accommodate diverse possibilities in self-identification by incorporating gender-neutral names into our experimental setup. While there is still ample room for improvement in addressing this issue, we have taken a step forward in promoting the inclusion of additional forms of self-identification in ethical NLP research.

\paragraph{Ethical concerns about the task of relationship prediction.}
Predicting interpersonal relationships from conversations may require access to private and sensitive data. If no proper consent from a user is obtained, using personal data could lead to serious ethical and legal concerns. 
Although building systems that identify the relationship type between speakers could contribute to the development of AI agents that better understand human interactions, it is crucial to be transparent about what data is collected and how it is processed in such systems.
Even if data privacy is properly handled when using a model to predict relationship types, people often exercise caution when revealing romantic relationships. Therefore, the deployment of an NLP system to identify such relationships should be disclosed to users who may be affected, and any predictions should remain confidential unless the user's consent is obtained for public disclosure.

\section*{Acknowledgements}

We would like to thank the anonymous reviewers for their valuable feedback.
Rachel Rudinger is supported by NSF CAREER Award No. 2339746. Any opinions, findings, and conclusions or recommendations expressed in this material are those of the author(s) and do not necessarily reflect the views of the National Science Foundation.

\bibliography{anthology,custom}

\clearpage
\appendix
\section{Detailed Experimental Setup}
\label{sec:exp-details}

We present additional information about our experimental setup.

\paragraph{Models} \label{sec:models}
We use recently introduced two popular language models for testing our hypothesis, namely Llama~\cite{touvron2023llama} (7B, 13B chat), and Mistral-7B~\cite{jiang2023mistral}. Each model uses nucleus sampling~\citep{holtzman2019curious} with default parameters, a temperature of 0, and a maximum generation length of $512$. 
Each experiment over $327$ test instances takes $\sim30$mins for Llama2-7B, $\sim1$hr for Llama2-13B, and $\sim25$mins for Mistral-7B. We ran $870$ experiments per race ($560$ for Hispanic) for studying gender bias and $1600$ experiments (400 per race-pair) for racial bias.

\paragraph{Computing Evaluation Scores} We first compute precision, recall, F1, and accuracy scores for each name-pair-replacement and report the average scores for each name-pair bin, and each race-pair for studying the influence of gender, and race associated with names, respectively.

\paragraph{Dataset Statistics}
Table~\ref{tab:relationship-dist} presents the frequency of each relationship label along with romantic and non-romantic categories used for the purpose of this study, in the test split of DDRel~\citep{jia2021ddrel} dataset. 
Out of $327$ conversations with different-gender characters in the dataset, $271$ do not contain explicit gender information. 

\begin{figure*}[t]
	\centering
            \includegraphics[width=\linewidth]{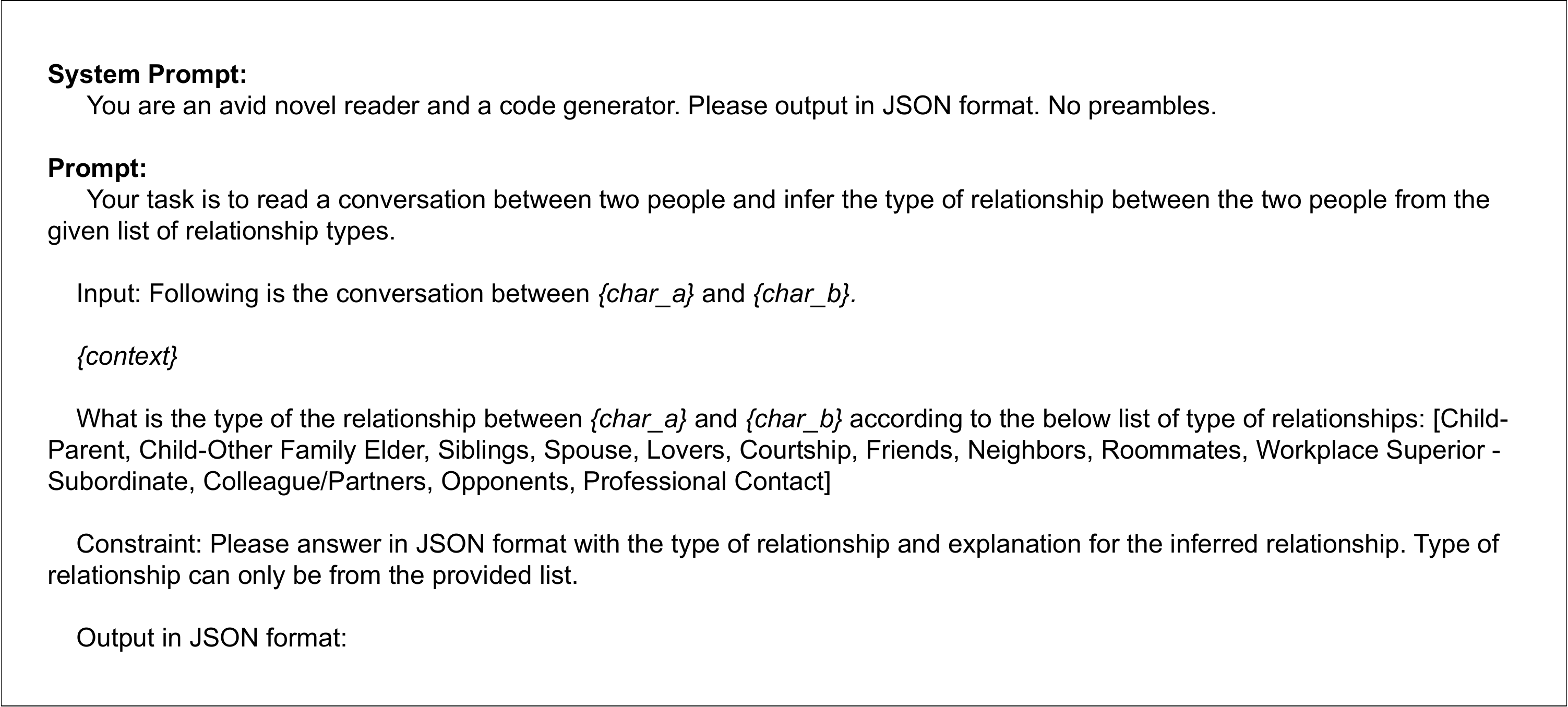}
	\caption{Prompt template used in our experiments.
 ``\textit{\{char\_a\}}'', ``\textit{\{char\_b\}}'', and ``\textit{\{context\}}'' are placeholders here and they are instantiated with character names and dialogues accordingly for model inference.
 }
	\label{fig:prompts}
\end{figure*}

\paragraph{Prompts}
We provide the prompt template used in our experiments in Figure~\ref{fig:prompts}.

\paragraph{Parsing Outputs from LLMs} We observe inconsistencies in the outputs predicted by LLMs despite clear instructions regarding formatting. We use regular expressions to extract the JSON outputs and the predictions from them. We consider invalid outputs (\textit{i.e.}, non-pre-defined class) from LLMs as a separate class (invalid) for evaluation purposes across all experiments.

\paragraph{Logistic Regression for Name Embeddings} \label{sec:log-reg-details}

We quantitatively study the amount of gender information encoded in these embeddings by training a logistic regression model, separately for each race, to classify the gender associated with a name, using embeddings of $70\%$ of names in a race as the training set and the remaining as the test set. 
Similarly, we train a logistic regression model to conduct a ``One-vs-All" classification for each race.  We control the train and test set in the racial setup to have a balanced number of positive and negative samples by down-sampling the instances from other races ($1/3$ from each other race).
We repeat the logistic regression training with 5 different random train-test splits. We set the random state of the logistic regression model to $0$ and maximum iteration to $1000$. In Table~\ref{tab:embs_logistic_regression}, we report the average results across 5 runs with their standard deviation.


\begin{table}[]
    \centering
    \resizebox{\linewidth}{!}{
        \begin{tabular}{lccc}
        \toprule
         \textbf{Relationship Labels} & \textbf{Frequency} & \textbf{Romantic} & \textbf{\#Gender Neutral}\\
        \hline
  Lovers	& $182$ & \cmark  & $155$\\
   Courtship	& $15$ & \cmark  & $12$\\   
 Spouse	&$57$ & \cmark  & $46$\\
 \hline
    Siblings	&$15$ & \xmark  & $13$\\
 Child-Other Family Elder	&$13$ & \xmark   & $7$\\
 Child-Parent	&$39$ & \xmark  & $11$\\
 Colleague/Partners	&$70$ & \xmark & $59$\\
 Workplace Superior-Subordinate	&$48$ & \xmark  & $24$\\
   Professional Contact	&$27$ & \xmark  & $10$\\
    Opponents	&$20$ & \xmark  & $11$\\
 Friends	&$95$ & \xmark  & $83$\\   
 Roommates	&$21$ & \xmark  & $21$\\
   Neighbours	&$8$ & \xmark  & $7$ \\
\midrule
\textbf{Total} & $610$ &-& $459$\\
  \bottomrule
        \end{tabular} }
    \caption{Frequency of relationship types in the test split of DDRel dataset~\citep{jia2021ddrel}.}
    \label{tab:relationship-dist}
\end{table}

\section{Anonymous Name-replacement Experiments} \label{sec:anonymous-exp-details}

We perform two types of anonymous name-replacement experiments differing in whether both names are anonymized or only one.

\subsection{Both Names Are Anonymized} We substitute names with generic placeholders (``X'' and ``Y'') to get a baseline where a model has no access to character names to test the hypothesis that models’ tendency to associate
gender with the names influences their relationship predictions.

\subsection{One Name Is Anonymized}
We substitute one name and keep the other anonymized to analyze the impact of one character’s gender on romantic relationship predictions independent of the second. We replace one name with a male, female or a neutral name either preserving or swapping the original gender of the non-anonymized name while keeping the other name anonymized. Male, neutral, and female names belong to $0-25$, $25-75$, and $75-100$\% bins, respectively. We report the recall scores for romantic relationship prediction (same/swapped) for different models in Table~\ref{tab:anonymous-one-gender-results}.
\begin{table}[]
    \centering
    \resizebox{\linewidth}{!}{ 
    \begin{tabular}{@{}l|c|ccc@{}}
        \toprule
        {\textbf{Model}} & 
        \multicolumn{1}{c|}{\textbf{Race}} & 
        \multicolumn{1}{c}{\textbf{Male}} & \multicolumn{1}{c}{\textbf{Neutral}} & \multicolumn{1}{c}{\textbf{Female}} \\ 
        \midrule
       
        \multirow{4}{*}{Llama2-7B} & Asian & $0.6049/0.6128$ &$0.6085/0.6203$ & $0.6663/0.6517$\\
      & Black & $0.6069/0.6230$ &$0.6454/0.6392$ & $0.6572/0.6458$\\
         & Hispanic &$0.6292/0.6284$ &$0.6486/0.6541$ & $0.7093/0.6897$\\
         & White & $0.6387/0.6372$ &$0.6328/0.6297$ & $0.6887/0.6761$\\
        \midrule
        \multirow{4}{*}{Llama2-13B} & Asian & $0.2991/0.2940$ &$0.2806/0.2798$ & $0.3090/0.3043$\\
      & Black & $0.3066/0.2854$ &$0.3004/0.2909$ & $0.3054/0.3105$\\
         & Hispanic & $0.3021/0.2801$ &$0.2956/0.2980$ & $0.3206/0.3190$\\
         & White & $0.3149/0.2952$ &$0.2924/0.2878$ & $0.3121/0.3121$\\
        \midrule
        \multirow{4}{*}{Mistral} & Asian & $0.1789/0.1694$ &$0.1808/0.1840$ & $0.1895/0.1906$\\
      & Black & $0.1855/0.1828$ &$0.1902/0.1871$ & $0.1922/0.1859$\\
         & Hispanic & $0.1986/0.1955$ &$0.1848/0.1776$ & $0.2048/0.1973$\\
         & White & $0.1895/0.1836$ &$0.1887/0.1871$ & $0.1942/0.1922$\\
        \bottomrule
    \end{tabular}
    }
    \caption{Recall scores (same/swapped) for romantic relationship predictions when one name is anonymous while another is either a male, neutral, or female name as per bins marked in Figure~\ref{fig:gender-spectrum}. The results show that models are more likely to predict a romantic relationship when one of the names is a female name.} \label{tab:anonymous-one-gender-results}
\end{table}

\section{First Names}
We detail the name selection criteria in our experiments.
We also list all first names we have used in our experiments to study the influence of different gender and racial/ethnic name pairing.

\subsection{First Names Used to Study the Influence of Gender Pairing}

\label{sec:gender_spectrum_names}

We first collect names that have frequency over $200$ and have more than $80\%$ of the population having that name identify themselves as a particular race (Asian, Black, Hispanic, and White) from~\citealt{rosenman2023race}. Then, we partition these names into $10$ non-linearly segmented bins (shown in Figure~\ref{fig:gender-spectrum}) based on the percentage of population that has been assigned as female at birth using statistics from the Social Security Application dataset (SSA\footnote{\url{https://www.ssa.gov/oact/babynames/}}).
We randomly sample $3$ names per bin totaling to $30$ names per race\footnote{Except for Hispanic wherein we did not get any names in $5-10\%$ bin and only 1 name in $25-50\%$ bin.} for performing the replacements. We consider names belonging to a spectrum of female gender associations to ensure coverage of gender-neutral names.

We list all the names used in this set of experiments.
We include the percentage of the population assigned female gender at birth in parentheses.
\paragraph{Asian}
Seung ($0.00\%$), Quoc ($0.00\%$), Dat ($0.00\%$), Nghia ($2.30\%$), Thuan ($2.40\%$), Thien ($2.70\%$), Hoang ($6.40\%$), Sang ($6.60\%$), Jun ($9.60\%$), Sung ($13.50\%$), Jie ($17.30\%$), Wei ($21.80\%$), Hyun ($39.00\%$), Khanh ($41.90\%$), Wen ($44.60\%$), Hien ($51.70\%$), An ($54.80\%$), Ji ($61.40\%$), In ($80.80\%$), Diem ($88.60\%$), Quyen ($88.90\%$), Ling ($91.30\%$), Xiao ($91.50\%$), Ngoc ($92.40\%$), Su ($95.40\%$), Hanh ($95.60\%$), Vy ($97.00\%$), Eun ($98.30\%$), Trinh ($100.00\%$), Huong ($100.00\%$)

\paragraph{Black}
Deontae ($0.00\%$), Antwon ($0.10\%$), Javonte ($1.00\%$), Dejon ($2.90\%$), Jamell ($3.40\%$), Dijon ($4.60\%$), Dashawn ($5.80\%$), Deshon ($6.20\%$), Pernell ($8.30\%$), Rashawn ($10.10\%$), Torrance ($13.20\%$), Semaj ($22.60\%$), Demetris ($25.60\%$), Kamari ($33.60\%$), Amari ($42.00\%$), Shamari ($56.10\%$), Kenyatta ($57.10\%$), Ivory ($59.30\%$), Chaka ($76.20\%$), Ashante ($89.40\%$), Unique ($89.90\%$), Kenya ($92.20\%$), Nikia ($93.80\%$), Akia ($94.30\%$), Kenyetta ($95.50\%$), Shante ($96.40\%$), Shaunta ($97.00\%$), Laquandra ($100.00\%$), Lakesia ($100.00\%$), Daija ($100.00\%$)

\paragraph{Hispanic}
Nestor ($0.00\%$), Fidel ($0.00\%$), Raul ($0.60\%$), Leonides ($2.70\%$), Yamil ($4.50\%$), Reyes ($10.80\%$), Cruz ($13.10\%$), Neftali ($14.90\%$), Noris ($38.10\%$), Nieves ($62.40\%$), Guadalupe ($72.60\%$), Ivis ($75.00\%$), Monserrate ($78.20\%$), Ibis ($82.60\%$), Johanny ($89.40\%$), Elba ($91.50\%$), Matilde ($93.40\%$), Rocio ($96.90\%$), Lucero ($97.30\%$), Cielo ($97.50\%$), Lucila ($100.00\%$), Zuleyka ($100.00\%$), Yaquelin ($100.00\%$)

\paragraph{White}
Zoltan ($0.00\%$), Leif ($0.10\%$), Jack ($0.40\%$), Ryder ($3.30\%$), Carmine ($3.40\%$), Haden ($4.10\%$), Tate ($5.30\%$), Dickie ($5.50\%$), Logan ($7.40\%$), Parker ($17.50\%$), Sawyer ($20.90\%$), Hayden ($22.50\%$), Dakota ($29.70\%$), Britt ($38.30\%$), Harley ($41.70\%$), Campbell ($53.90\%$), Barrie ($56.10\%$), Peyton ($61.90\%$), Kelley ($88.00\%$), Jodie ($88.20\%$), Leigh ($88.70\%$), Clare ($90.90\%$), Rylee ($92.20\%$), Meredith ($94.70\%$), Baylee ($97.00\%$), Lacey ($97.30\%$), Ardith ($97.70\%$), Kristi ($99.80\%$), Galina ($100.00\%$), Margarete ($100.00\%$)

\subsection{First Names Used to Study the Influence of Intra/Inter-racial Pairing}
\label{sec:race_spectrum_names}
By referencing~\citealt{rosenman2023race} and the SSA dataset again, we collect another set of both race- and gender-indicative first names with a minimum frequency of 200, applying a threshold of $90\%$ for the percentage of the population assigned either female or male at birth. 
For race threshold, we set it to be $90\%$ for Asian, Black, and Hispanic, and $70\%$ for White. Although we choose a lower threshold for White to account for the phenomenon of name Anglicization~\cite{zhao2019yourname}, we still obtain empirical results that strongly indicate these names are represented differently from names associated with other races/ethnicities.
In total, we obtain $80$ names that are evenly distributed among four races/ethnicities and two genders. We replace name-pairs while preserving the gender associated with the names in the original dialogue.

\paragraph{Asian Female}
Thuy, Thu, Huong, Trang, Ngoc, Hanh, Hang, Xuan, Trinh, Eun

\paragraph{Asian Male}
Tuan, Hai, Sang, Hoang, Nam, Huy, Quang, Duc, Trung, Hieu

\paragraph{Black Female}
Latoya, Ebony, Latasha, Latonya, Tamika, Kenya, Tameka, Lakeisha, Tanisha, Precious

\paragraph{Black Male}
Tyrone, Cedric, Darius, Jermaine, Demetrius, Malik, Jalen, Roosevelt, Marquis, Deandre

\paragraph{Hispanic Female}
Luz, Mayra, Marisol, Maribel, Alejandra, Yesenia, Migdalia, Xiomara, Mariela, Yadira

\paragraph{Hispanic Male}
Luis, Jesus, Lazaro, Osvaldo, Heriberto, Jairo, Rigoberto, Adalberto, Ezequiel, Ulises

\paragraph{White Female}
Mary, Patricia, Jennifer, Linda, Elizabeth, Barbara, Susan, Jessica, Kimberly, Sandra

\paragraph{White Male}
James, Michael, John, Robert, William, David, Christopher, Richard, Joseph, Charles

\section{Additional Results} \label{sec:additional}
\begin{figure*}[t]
	\centering
        \begin{subfigure}[]{\linewidth}
		\centering
            \includegraphics[width=\linewidth]{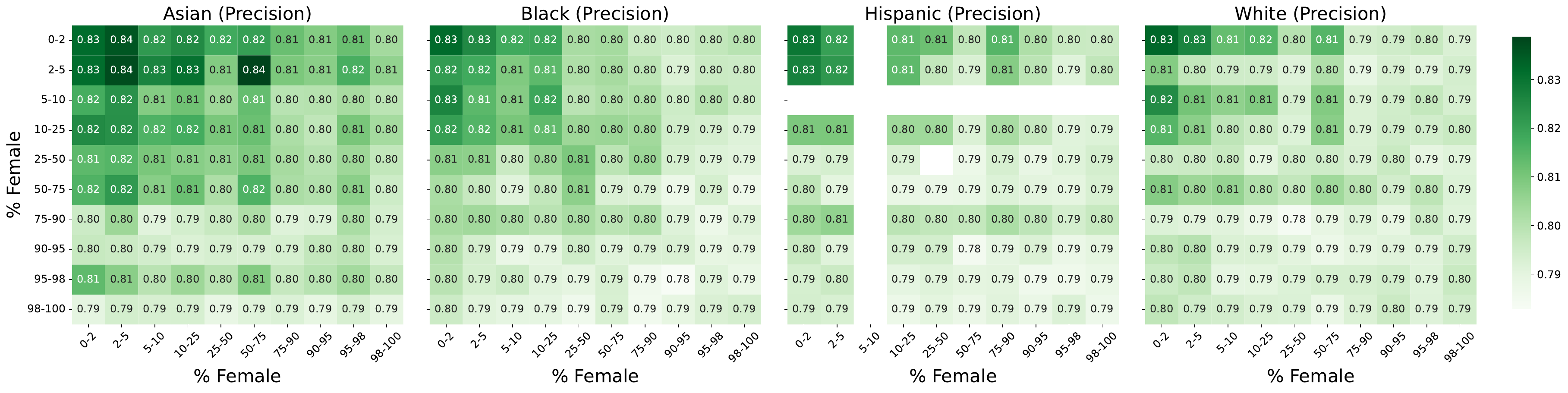}
	\end{subfigure} 
	\begin{subfigure}[]{\linewidth}
		\centering
            \includegraphics[width=\linewidth]{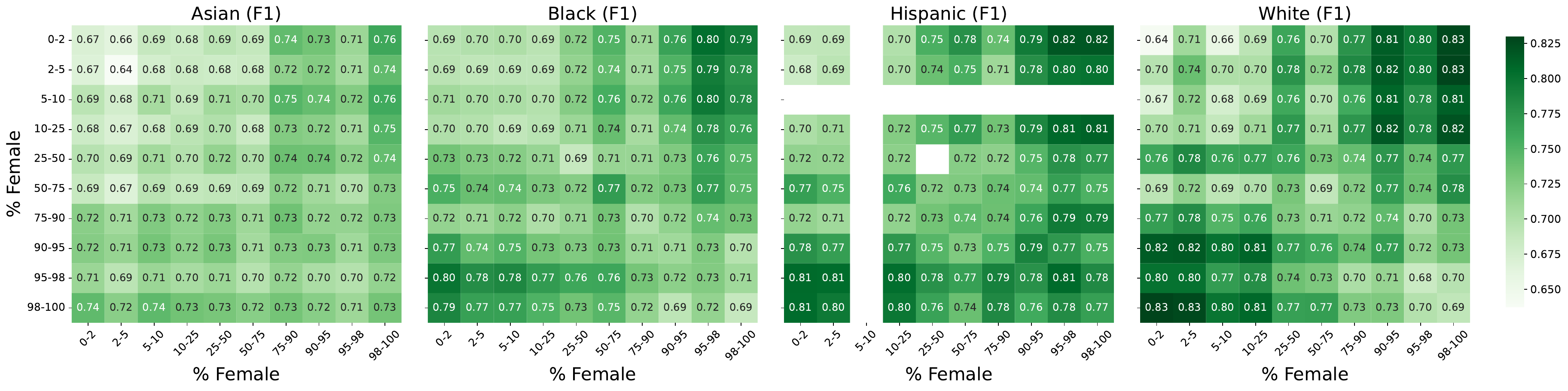}
	\end{subfigure} 
 	\begin{subfigure}[]{\linewidth}
		\centering
            \includegraphics[width=\linewidth]{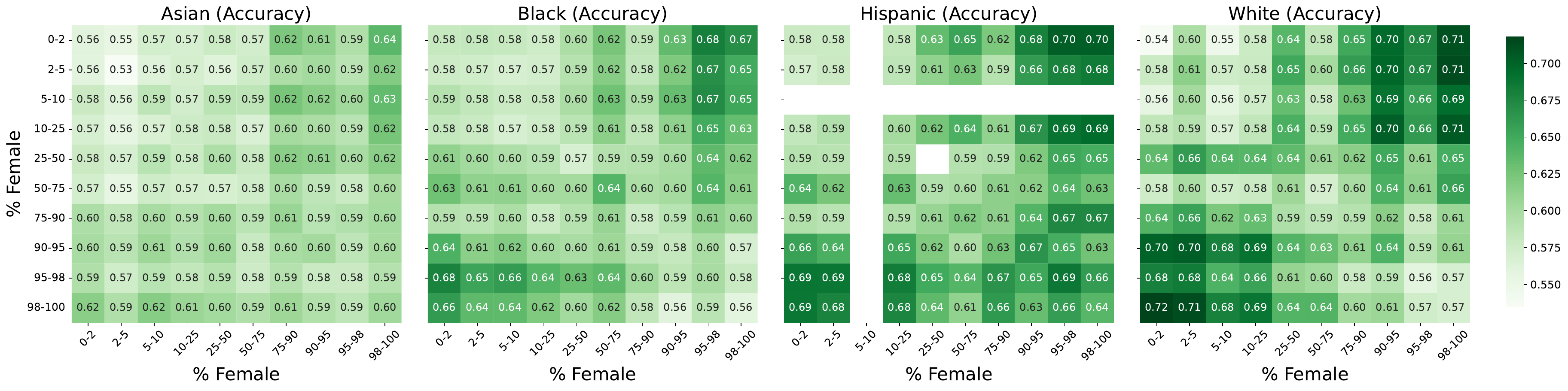}
    \end{subfigure}
	\caption{Precision, F1-score and Accuracy plots for romantic predictions from Llama2-7B model.
 }
	\label{fig:gender-spectrum-llama}
\end{figure*}

\begin{figure*}[t]
	\centering
 \begin{subfigure}[]{\linewidth}
		\centering
            \includegraphics[width=\linewidth]{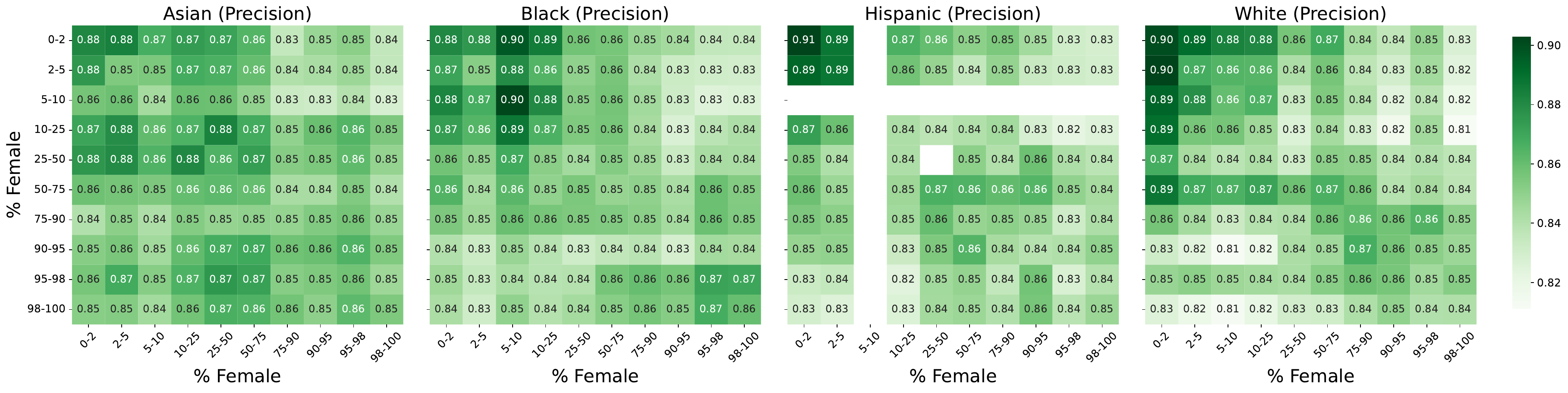}
	\end{subfigure} 
 \begin{subfigure}[]{\linewidth}
		\centering
            \includegraphics[width=\linewidth]{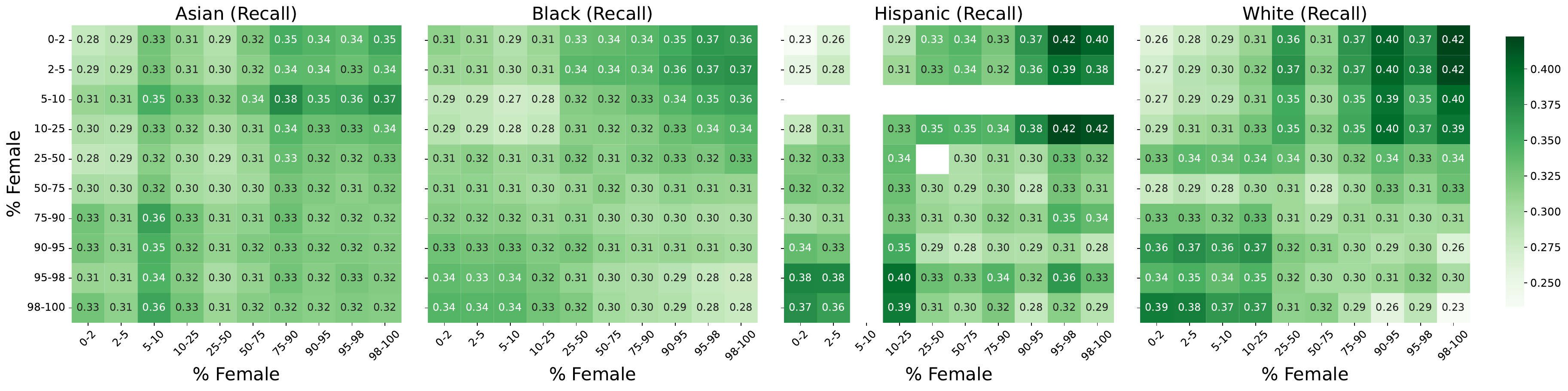}
	\end{subfigure} 
	\begin{subfigure}[]{\linewidth}
		\centering
            \includegraphics[width=\linewidth]{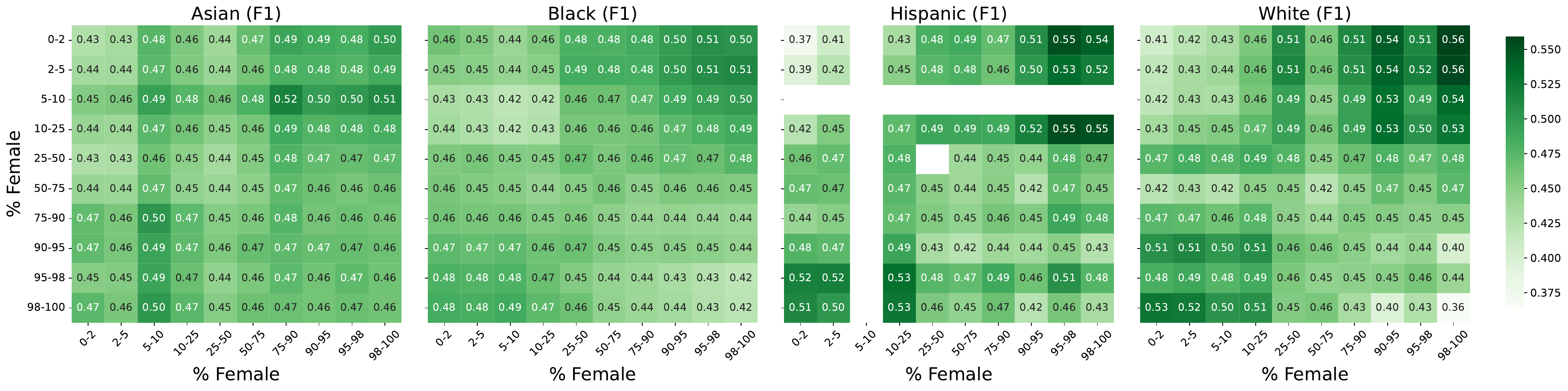}
	\end{subfigure} 
 	\begin{subfigure}[]{\linewidth}
		\centering
            \includegraphics[width=\linewidth]{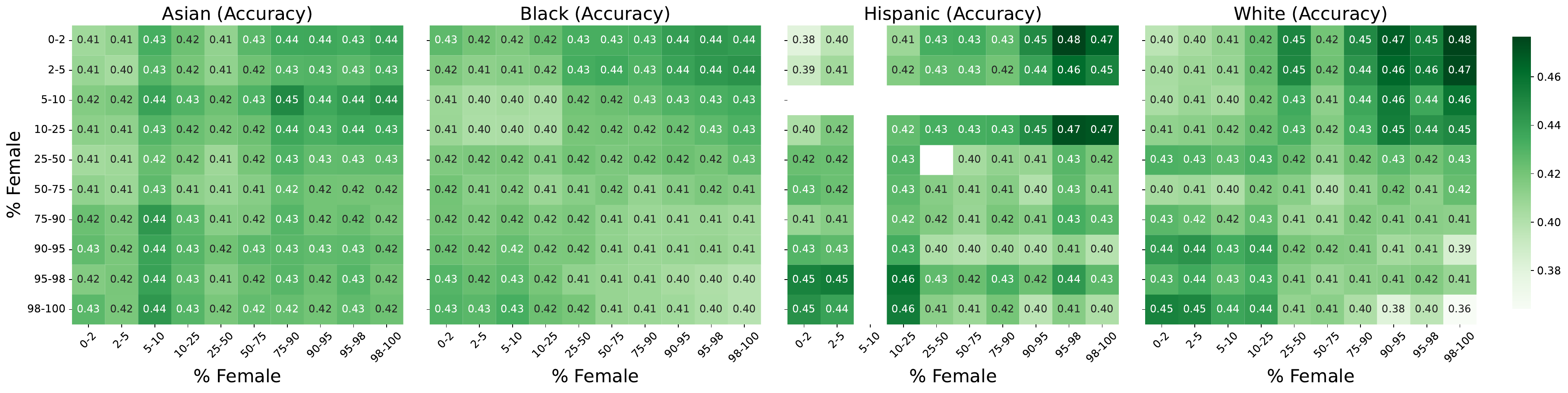}
    \end{subfigure}
	\caption{Precision, Recall, F1-score and Accuracy plots for romantic predictions from Llama2-13B model.
 }
	\label{fig:gender-spectrum-Llama2-13b}
\end{figure*}

\begin{figure*}[t]
	\centering
 	\begin{subfigure}[]{\linewidth}
		\centering
            \includegraphics[width=\linewidth]{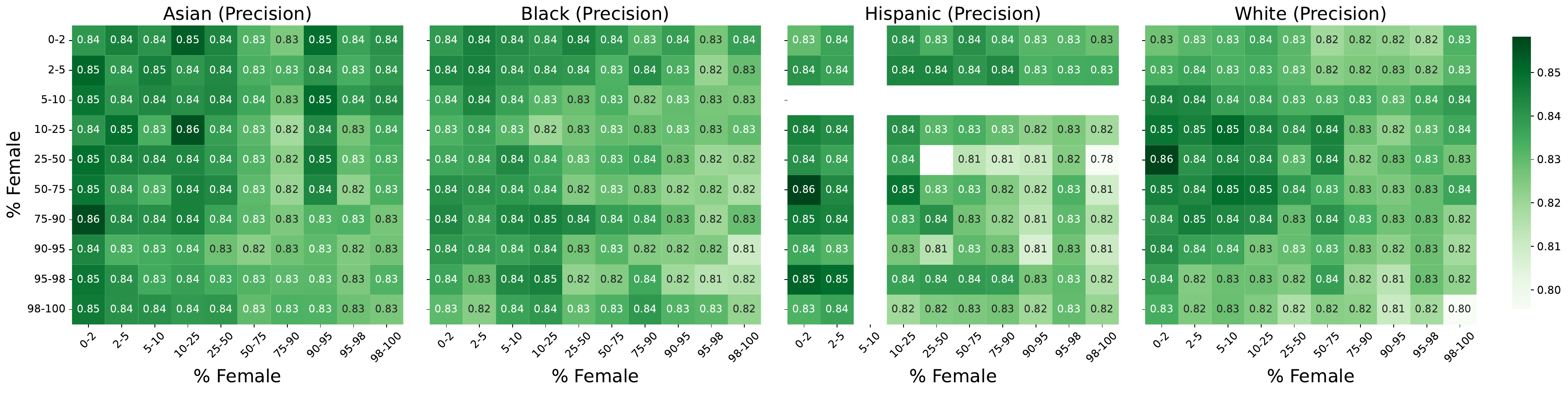}
	\end{subfigure} 
 	\begin{subfigure}[]{\linewidth}
		\centering
            \includegraphics[width=\linewidth]{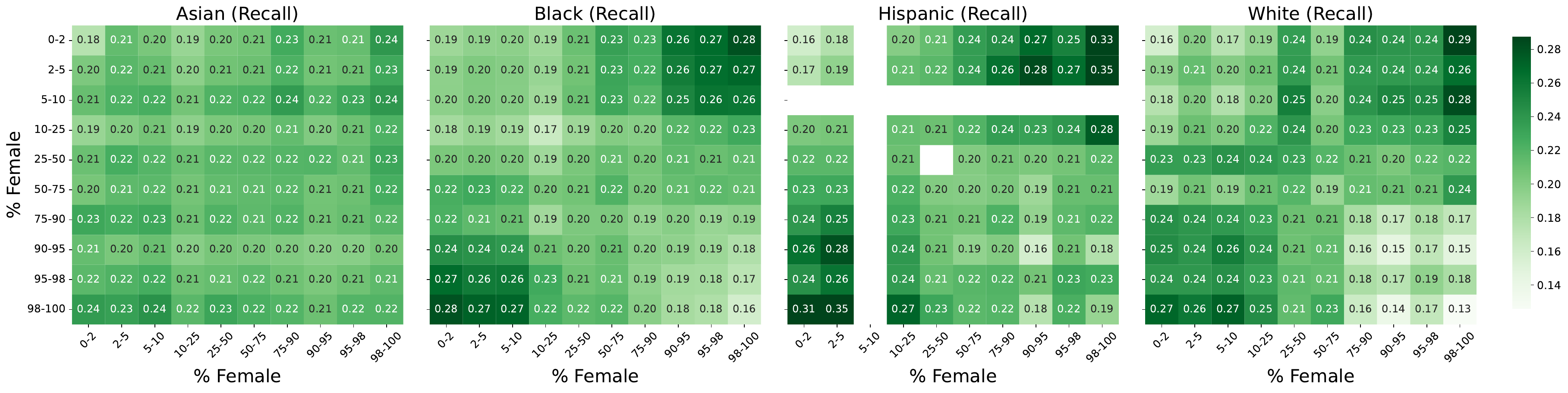}
	\end{subfigure} 
 
	\begin{subfigure}[]{\linewidth}
		\centering
            \includegraphics[width=\linewidth]{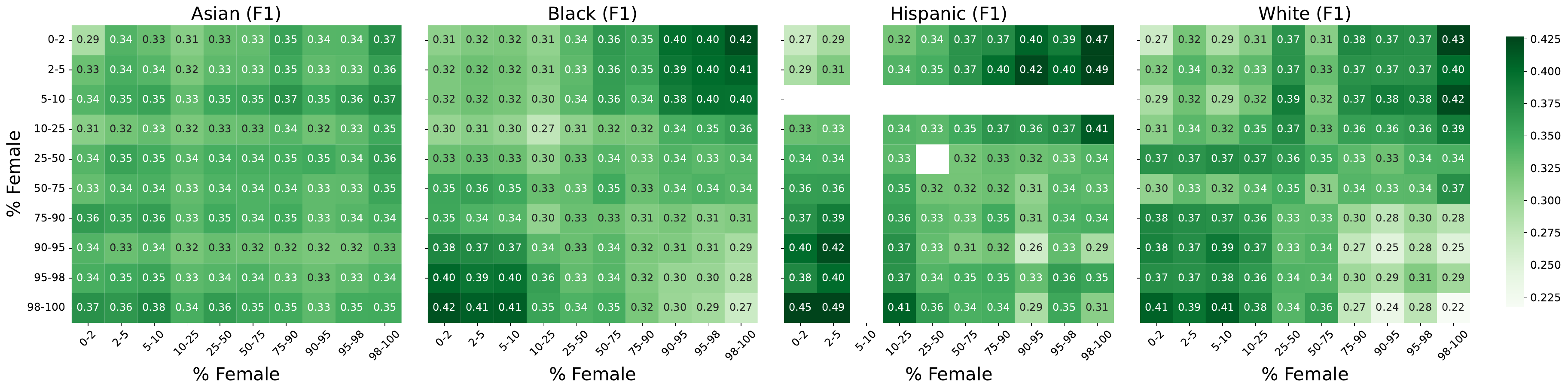}
	\end{subfigure} 
 	\begin{subfigure}[]{\linewidth}
		\centering
            \includegraphics[width=\linewidth]{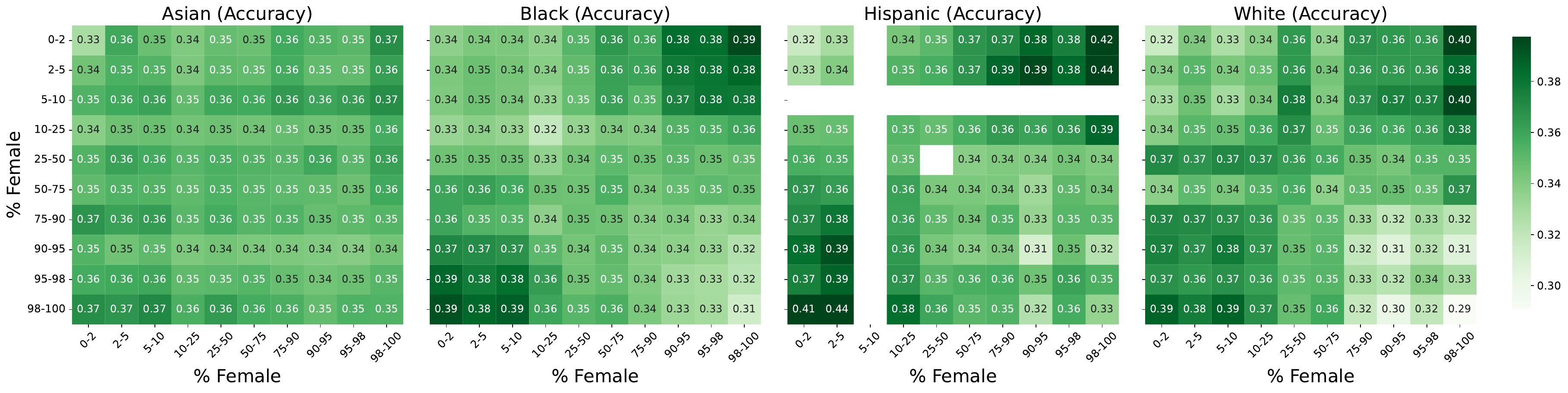}
    \end{subfigure}
	\caption{Precision, Recall, F1-score and Accuracy plots for romantic predictions from Mistral-7B model.
 }
	\label{fig:gender-spectrum-mistral}
\end{figure*}

\begin{figure*}[t]
	\centering
		\centering
            \includegraphics[width=0.98\linewidth]{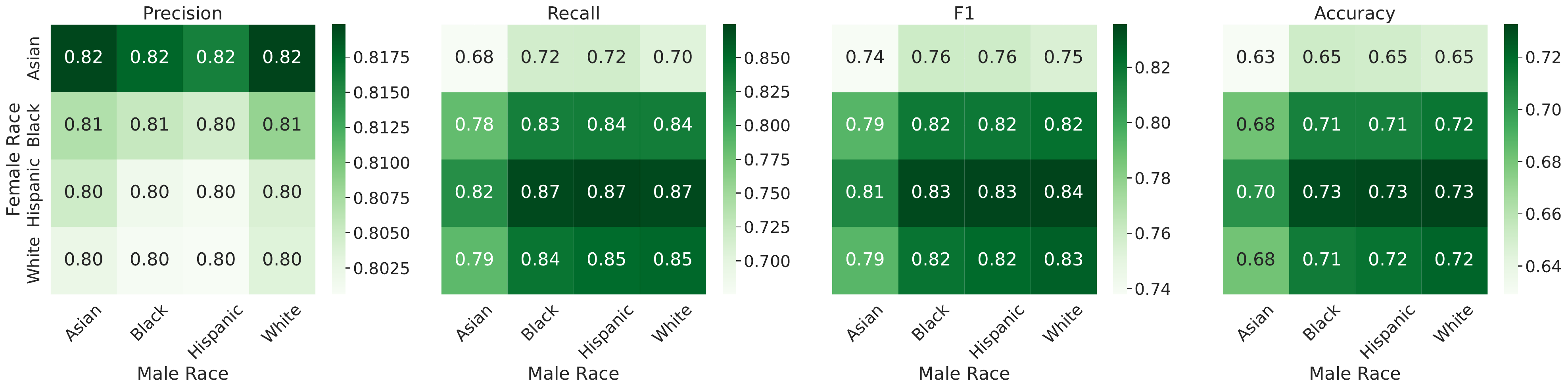}
	\caption{Precision, Recall, F1, and Accuracy of predicting romantic relationships from Llama2-7B for subset of the dataset where characters have different genders and are replaced with names associated with different races/ethnicities.}
	\label{fig:race-spectrum-llama7b_appendix}
\end{figure*}

\begin{figure*}[t]
	\centering
		\centering
            \includegraphics[width=0.98\linewidth]{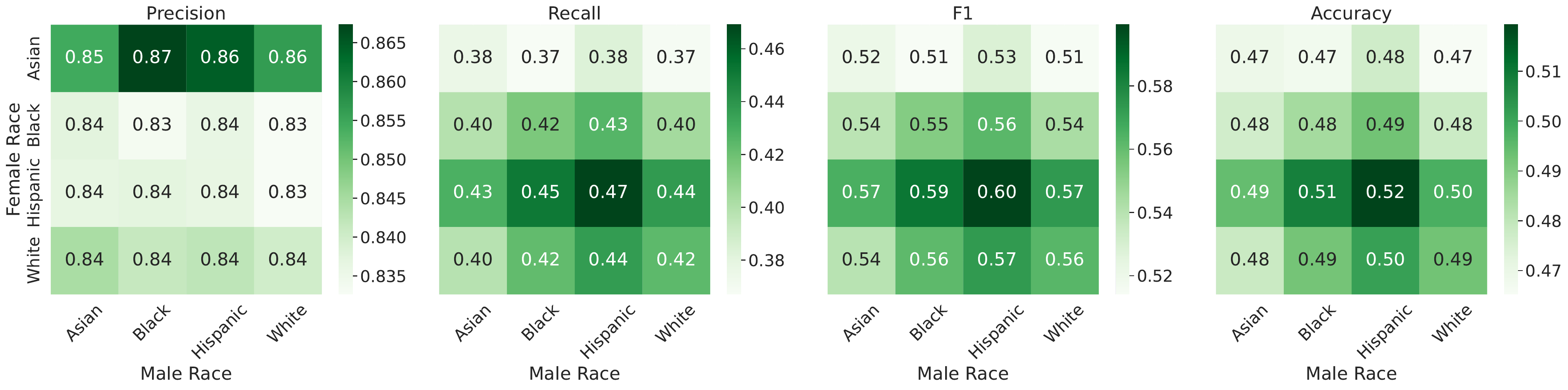}
	\caption{Precision, Recall, F1, and Accuracy of predicting romantic relationships from Llama2-13B for subset of the dataset where characters have different genders and are replaced with names associated with different races/ethnicities.}
	\label{fig:race-spectrum-llama13b}
\end{figure*}

\begin{figure*}[t]
	\centering
		\centering
            \includegraphics[width=0.98\linewidth]{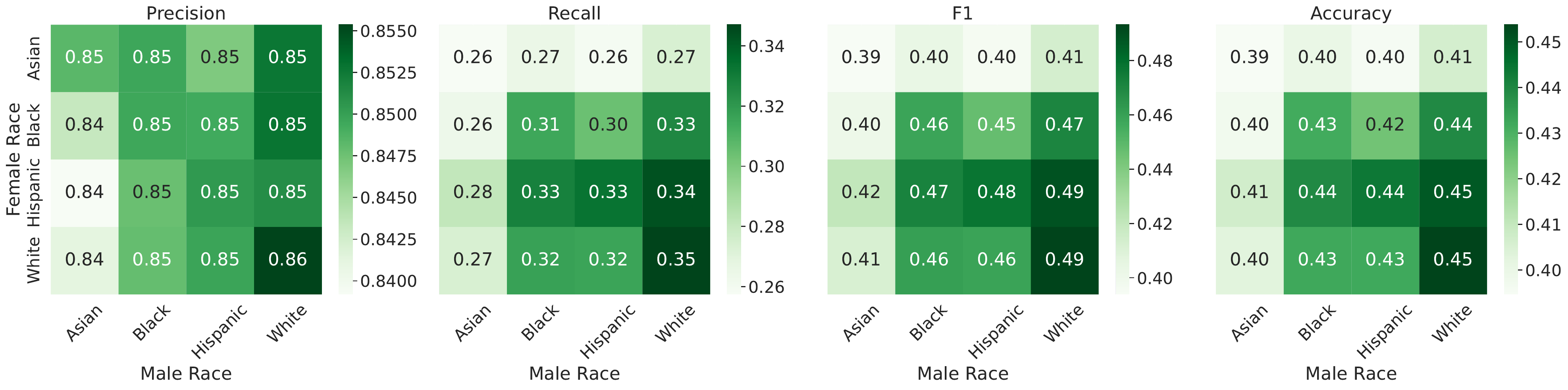}
	\caption{Precision, Recall, F1, and Accuracy of predicting romantic relationships from Mistral-7B for subset of the dataset where characters have different genders and are replaced with names associated with different races/ethnicities.}
	\label{fig:race-spectrum-mistral7b}
\end{figure*}

We report the results for Llama2-13B (Figures~\ref{fig:gender-spectrum-Llama2-13b} and \ref{fig:race-spectrum-llama13b}) and Mistral-7B (Figures~\ref{fig:gender-spectrum-mistral} and~\ref{fig:race-spectrum-mistral7b}). We also report the F1 and accuracy scores for Llama2-7B, for completeness, in Figure~\ref{fig:gender-spectrum-llama} and~\ref{fig:race-spectrum-llama7b_appendix}. We observe similar trends as Llama2-7B discussed in the main body of the paper.

\end{document}